\documentclass[a4paper]{svproc}

\usepackage{authblk}
\usepackage{cite}
\usepackage{amsmath}
\usepackage{amsfonts}
\usepackage{bm}
\usepackage{graphicx}
\usepackage{color}
\usepackage{units}
\usepackage{verbatim}
\usepackage{pgfplots}
\pgfplotsset{compat=newest}
\usetikzlibrary{external}
\usepackage{siunitx}
\usepackage{silence}

\begin{document}

\title{
\bf{A Recurrent Neural Network Approach to \\Roll Estimation for Needle Steering}
}

\author{
    Maxwell~Emerson\inst{1}, %\orcID{0000-0003-2491-1979},
    James~M.~Ferguson\inst{1},
    Tayfun~Efe~Ertop\inst{1},
    Margaret~Rox\inst{1},
    Josephine~Granna\inst{1},
    Michael~Lester\inst{2},
    Fabien~Maldonado\inst{2},
    Erin~A.~Gillaspie\inst{2},
    Ron~Alterovitz\inst{3},
    Robert~J.~Webster~III\inst{1},
    \and Alan~Kuntz\inst{4}
}

\institute{
    Department of Mechanical Engineering, Vanderbilt University
    \and Department of Medicine and Thoracic Surgery, \\Vanderbilt University Medical Center
    \and Department of Computer Science, University of North Carolina at Chapel Hill 
    \and Robotics Center and School of Computing, University of Utah
}

\titlerunning{RNN Approach to Roll Estimation for Needle Steering}
\authorrunning{Emerson \emph{et. al.}}
\maketitle

\vspace*{-0.5cm}

\section{Motivation}
\begin{figure}[t]
    \centering
    \includegraphics[width=\textwidth]{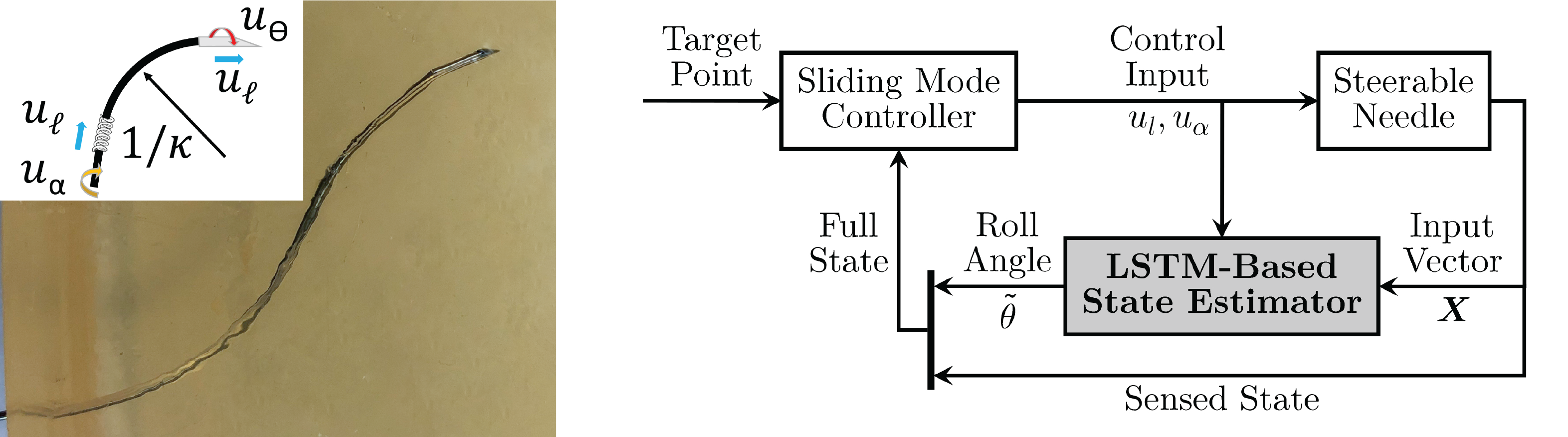}
    \caption{\textit{Left:} Steerable needle inserted into a gelatin tissue simulant. \textit{Left Inset:} Kinematic diagram of the steerable needle model with un-modeled torsional compliance. This results in rotational lag between the base of the needle and tip of the needle. \textit{Right:} Block diagram showing our LSTM-based estimator in the control loop.
    }
    \label{fig:kin_diagram}
    \vspace*{-0.5cm}
\end{figure}

\vspace*{-0.2cm
}
Steerable needles are a promising technology for delivering targeted therapies in the body in a minimally invasive fashion via controlled, actively steered insertions.
These robotically actuated needles usually leverage an asymmetric tip~\cite{Webster2006} to take curved paths through the body, avoiding anatomical obstacles and honing in on a target (see Fig.~\ref{fig:kin_diagram}, left). 
Methods such as duty cycling and sliding mode control enable safe, accurate, and automatic controlled steering of these needles to anatomical targets or along predetermined trajectories in the body~\cite{Rucker2013, Minhas2007}.
These controllers require knowledge of the full 6 degrees of freedom (DOF) pose of the steerable needle's tip as it is steered through the body.
To acquire this information for feedback during control, electromagnetic sensors can be embedded in the tip of the needle.
However, these sensors typically fill the internal working channel of the needle, precluding the use of the needle for therapy delivery.
External sensors, such as ultrasound and bi-plane fluoroscopy (a type of continuous X-ray) can sense the needle tip's position and heading (5 DOF), but are not able to sense its axial orientation (roll angle)~\cite{Swensen2014}.
Without full 6-DOF state measurement, an alternative method is needed to estimate the full state of the needle tip for effective steering.

Model-based observer methods have been developed to estimate the orientation of the needle during steering~\cite{reed2009modeling, kallem2009image, fallahi2016partial, Wood2010}.
These methods work well when the system behaves similarly to the modeled system they rely on. However, when unmodeled effects dominate the system dynamics these methods can perform poorly.
For flexible needles, effects such as unpredictable friction forces due to tissue interactions and long needle lengths create nonlinear torsional dynamics that are difficult to accurately model \cite{kallem2009image, reed2009modeling, Swensen2014}. 

By contrast, model-free, data-driven approaches have recently been of great research interest in state estimation during control in other domains \cite{Kuutti2019}.
However the investigation of such methods has so-far been relatively limited in needle steering, e.g., to predicting needle deflection behavior for set insertion depths~\cite{Avila-Carrasco2020}.

In this work, we overcome the limitations present in model-based observers for needle steering by presenting an estimator that \emph{learns} the behavior of the needle from a set of training insertions in a tissue phantom. 
Our method is capable of accurately estimating the needle's roll angle during new insertions and in tissues that are different from those that it trained on.

Using a model-free representation that learns the nonlinear effects from the training data, we achieve accurate state estimation that abstracts to multiple types of tissue for a system that is subject to significant modeling errors.

Here, we propose a Long-Short-Term Memory (LSTM)-based recurrent neural network (RNN)~\cite{Greff2017} to estimate the needle's roll angle during steering.
Our LSTM-based network takes as input the sensed, partial state of the needle tip and recurrently estimates the needle's unsensed roll angle at each time step.
We use the network to estimate the needle state in a sliding mode controller and demonstrate highly accurate steering in multiple mediums---including \emph{ex vivo} ovine brain and \emph{ex vivo} porcine inflated lung---significantly outperforming a traditional Extended Kalman Filter (EKF) estimation method reliant on a kinematic model that does not account for torsional effects in the system.

The key contributions of this paper are (i) an accurate, model-free, learning-based method for steerable needle roll angle estimation and (ii) the integration of the method into sliding mode control for highly accurate steering in multiple mediums, including \emph{ex vivo} tissue that the model was not trained on.
As such, our learned method can be trained in advance in gelatin using an internal sensor which can then be removed from the needle prior to the needle's use during clinical deployment in a patient.
The needle can then be controlled using external partial sensing and our learned method.
In this way, our learning-based method overcomes a key limitation in needle steering, namely the requirement for bulky 6-DOF sensors embedded in the needle itself during clinical deployment.
This opens the door for external needle state sensing, enabling accurate tip orientation estimation (and subsequently safe and accurate needle steering) in a way that does not interfere with the needle's ability to deliver therapy to the patient.

\vspace*{-0.2cm}
\section{Technical Approach}
\vspace*{-0.2cm}

We use a kinematic, non-holonomic needle model \cite{Webster2006} to define the state of a bevel-tip steerable needle as it moves through tissue.
In this model, the needle's behavior is parameterized by its forward motion, transmitted from actuation at its base; the plane in which its bevel lies and in which it curves, changed by rotating the needle at its base; and the curvature of the arc ($\kappa$) the needle takes as it is inserted. 
The control inputs of this model are $u_\ell$ and $u_{\theta}$, needle tip insertion velocity and angular velocity, respectively, as shown in Fig.~\ref{fig:kin_diagram}, left inset.
Most models assume the needle is infinitely rigid in torsion such that $u_{\alpha}$, the angular control velocity applied at the needle's base, is perfectly and immediately applied to the needle's tip. 
In reality, there is a lag in transmission of the rotational velocity applied at the actuator to the needle tip, i.e. $u_\alpha \neq u_\theta$, an effect that is particularly pronounced for long and/or highly flexible needles.
Our method overcomes this limitation by learning to estimate the tip angle $\theta$ so that the controller can accurately steer the needle.

To do so, we propose a learning-based recurrent neural network with the following layers: (i) input sequence layer (5 units), (ii) LSTM layer (30 units), (iii) fully-connected layer (30 units), and (iv) output regression layer (2 units). 

Our network takes as input the vector $\bm{X} = [\bm{\hat{p}} \hspace{2mm} \bm{\hat{\eta}} \hspace{2mm} \sin{\alpha} \hspace{2mm} \cos{\alpha}]^T$,  where $\bm{\hat{p}} = (\hat{x} \hspace{2mm} \hat{y} \hspace{2mm} \hat{z})$ is the 3-DOF sensor position, isotropically scaled by a predefined maximum workspace component (e.g., the needle's maximum insertion length in tissue) using min-max feature scaling. The 2-DOF sensor axis measurement of the needle tip is given by $\bm{\hat{\eta}} = (\eta_x \hspace{2mm} \eta_y \hspace{2mm} \eta_z)$, representing the needle's heading (i.e., its orientation but without the roll angle). The rotational actuator position at the base of the needle is given by $\alpha = \int u_\alpha dt$. The network then outputs the vector $\bm{Y} = [\sin{\tilde{\theta}} \hspace{2mm} \cos{\tilde{\theta}}]^T$ where $\tilde{\theta}$ is the estimated roll angle.
We parameterize the input and output roll angles via $\sin$ and $\cos$, as these continuous representations nicely bound the variables from $[-1,1]$.

To utilize the network's estimated rotational angle, $\tilde{\theta}$, we integrate the network into a sliding mode controller~\cite{Rucker2013}, as shown in Fig.~\ref{fig:kin_diagram}, right.
$\tilde{\theta}$ is then applied to the needle's sensed heading $\bm{\hat{\eta}}$ at each time step of the control loop which, in combination with the sensed position $\bm{\hat{p}}$, enables the controller to accurately steer the needle via the estimated knowledge of the needle tip's full orientation.

\vspace*{-0.3cm}
\vspace*{-0.15cm}

\section{Experiments and Results}
\vspace*{-0.2cm}

\begin{figure}[t]
    \centering
    \includegraphics[width=\textwidth,height=4cm]{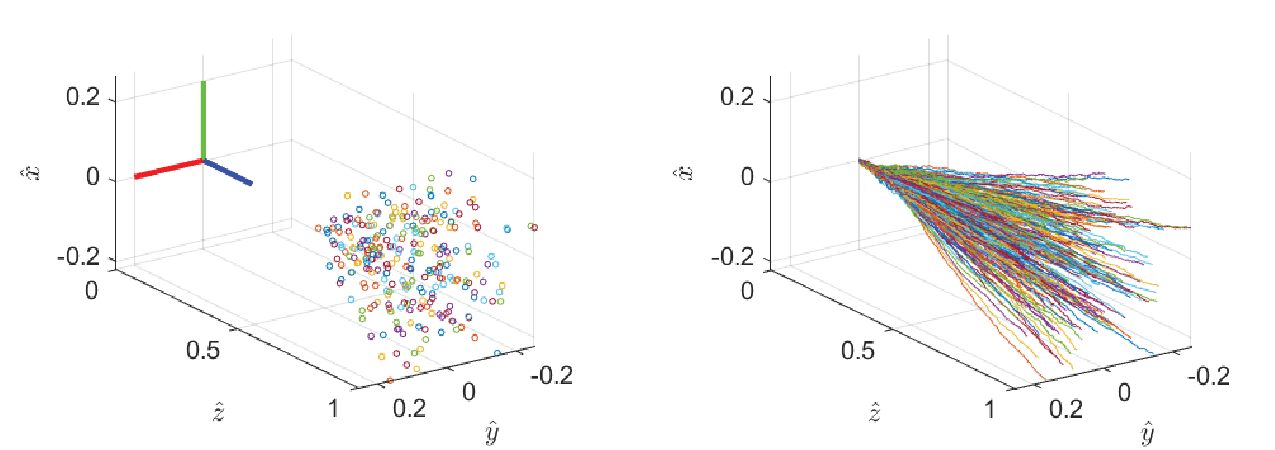}
    \vspace*{-0.8cm}
    \caption{Experimental dataset of 270 insertions in gelatin for network training. \textit{Left:} The target points. \textit{Right:} The trajectories taken to the target points. Data is isotropically scaled based on $z_{max}$, which we choose to be an insertion length of \unit[75]{mm}. 
    }
    \label{fig:normed_data}
    \vspace*{-0.6cm}
\end{figure}

\textbf{Data Collection and Network Training:}
We implemented and evaluated our learning-based method on a robotic needle steering system previously presented in~\cite{Amack2019}, designed to perform lung tumor biopsy through a bronchoscope.
The steerable needle used was manufactured out of superelastic Nitinol measuring $\unit[1.24]{mm}$ OD, $\unit[1.0]{mm}$ ID, $\unit[1.3]{m}$ long (EuroFlex GmbH), fabricated using the method described in \cite{RoxAccess2020}, and deployed through a clinical bronchoscope (Ambu USA). 
To collect the training data for our method, in the form of time series sequences of paired input and output vectors $\bm{X}$ and $\bm{Y}$, we performed targeting insertions using a sliding mode controller \cite{Rucker2013} with a 6-DOF electromagnetic (EM) tracker embedded in the needle tip (Aurora NDI, Inc.).
We collected a dataset of 270 insertions in a tissue simulant of 10\% gelatin (a tissue phantom frequently used in the needle steering literature)~\cite{Swaney2013}. 
Each insertion targeted a point sampled uniformly at random within the needle's reachable workspace defined by a cone with bounding curvature of $\unit[200]{mm^{-1}}$, and along an insertion interval of $\unit[40-75]{mm}$ (Fig.~\ref{fig:normed_data}, left).
Every insertion used the controller parameters $\lambda_1 = \unit[5]{mm/sec}$ and $\lambda_2 = \unit[2\pi]{rad/sec}$, a \unit[40]{Hz} controller rate, and achieved less than $\unit[1]{mm}$ targeting error (Fig.~\ref{fig:normed_data}, right).

We then normalized and partitioned the data into two subsets: training (240 insertions) and validation (30 insertions).
The network was trained using a root-mean-squared error loss function on an error defined over all timesteps in all trajectories in the training dataset. 
It was trained on an Intel i9-7900X 3.3GHz 10-core CPU with an NVIDIA Quadro P4000 GPU using ADAM optimization~\cite{Kingma2015} and dropout regularization~\cite{GalY.Ghahramani2016} to prevent over-fitting.
The trained network achieved a final RMSE of 0.0457 on the validation set.

\textbf{Online Estimation and Control in Gelatin and Ovine Brain:}
After the network was trained, we integrated it into the sliding mode controller to evaluate the system's ability to leverage our method to steer accurately to targets.
We performed 30 new insertions in 10\% gelatin to evaluate its performance in the tissue phantom in which it was trained and 10 new insertions in ovine brain preserved in 4\% Formalin (Carolina Biological Supply, Inc.) to evaluate its performance in biological tissue, a different medium and one in which it was \emph{not} trained.
The target points in both mediums were sampled uniformly at random from a similar volume as described in the training sets.
At each time step our method estimated the orientation angle $\theta$, which was combined with the sensed needle axis and position, forming the full state vector used for sliding mode control.

For comparison to a state-of-the-art method, we implemented an Extended Kalman Filter~\cite{choset2005principles}, a model-based observer relying on the non-holonomic kinematic needle model with no model of the torsional dynamics.
The EKF had knowledge of the same measured 5-DOF position and axis of the sensorized needle tip and the control input velocities applied at the actuators.
The EKF used the sensed information and kinematic needle model to estimate the state of the needle throughout the insertion, its estimated state being incorporated into the same sliding mode controller as our method. 

\begin{figure}[t]
    \centering
    \includegraphics[width=\textwidth]{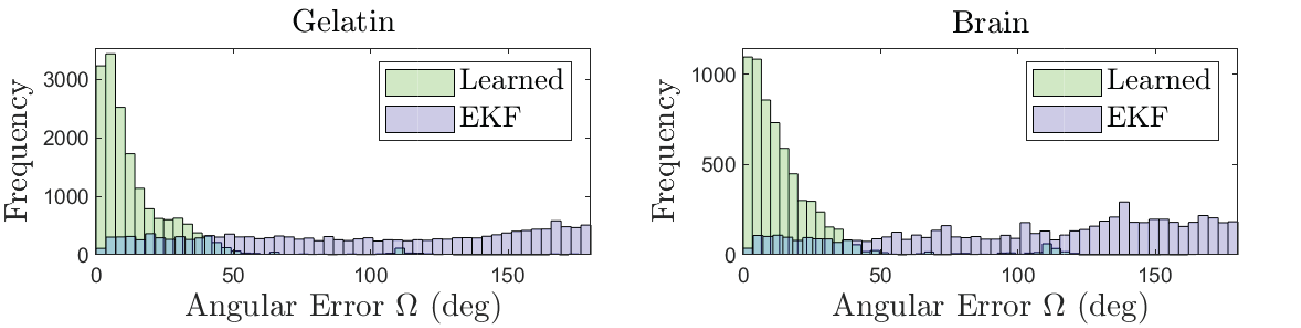}
    \vspace*{-0.8cm}
    \caption{Histograms of angular error over all insertions in gelatin and brain.}
    \label{fig:histograms_angular_error}
\end{figure}

In Fig.~\ref{fig:histograms_angular_error}, we show histograms of the angular error compared with the ground truth measured by the sensor in the needle's tip for each time step over all insertions. The angular error is defined as $\Omega = \arccos{((tr(\bm{R})-1)/2)}$, where $\bm{R}$ is the difference rotation relating the estimated orientation to the ground truth orientation.
Our method has error values distributed much closer to zero than the EKF method, indicating superior estimation through each of the insertions.

\begin{figure}[t]
    \centering
    \includegraphics[width=\textwidth]{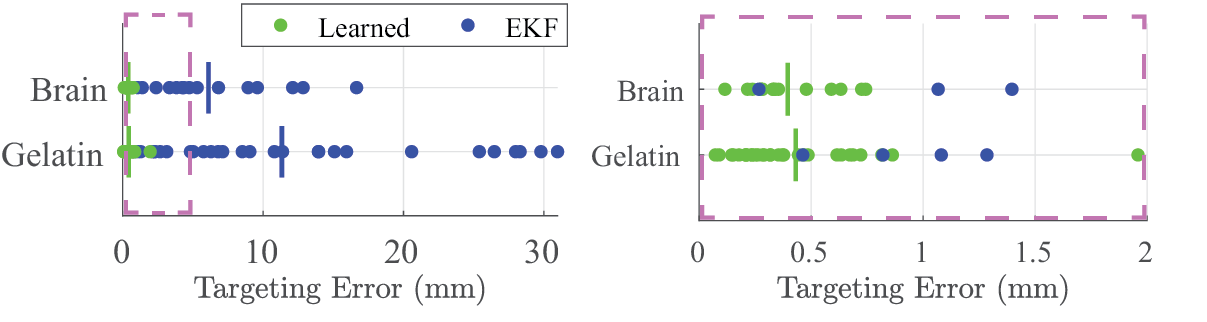}
    \vspace*{-0.8cm}
    \caption{Targeting errors in gelatin and ovine brain. \textit{Left:} Targeting errors from 10 insertions in gelatin and 5 insertions in brain  for our method (green) and the EKF (blue), with the straight lines depicting the average error for each. \textit{Right:} An expanded view of the 0 to 2 mm range of the left figure.
    }
    \label{fig:target_errors}
     \vspace*{-0.6cm}
\end{figure}

To demonstrate our method's ability to control the needle to its intended target, in Fig.~\ref{fig:target_errors} we show targeting errors (the Euclidean distance between the final needle tip position and the intended target) for each method in each medium.
Our method achieved mean targeting errors of \unit[0.43]{mm} in gelatin and \unit[0.40]{mm} in brain tissue, while the EKF method achieved mean targeting errors of \unit[11.34]{mm} in gelatin and \unit[6.12]{mm} in brain tissue.

\begin{figure*}[t]
    \centering
    \includegraphics[width=\textwidth]{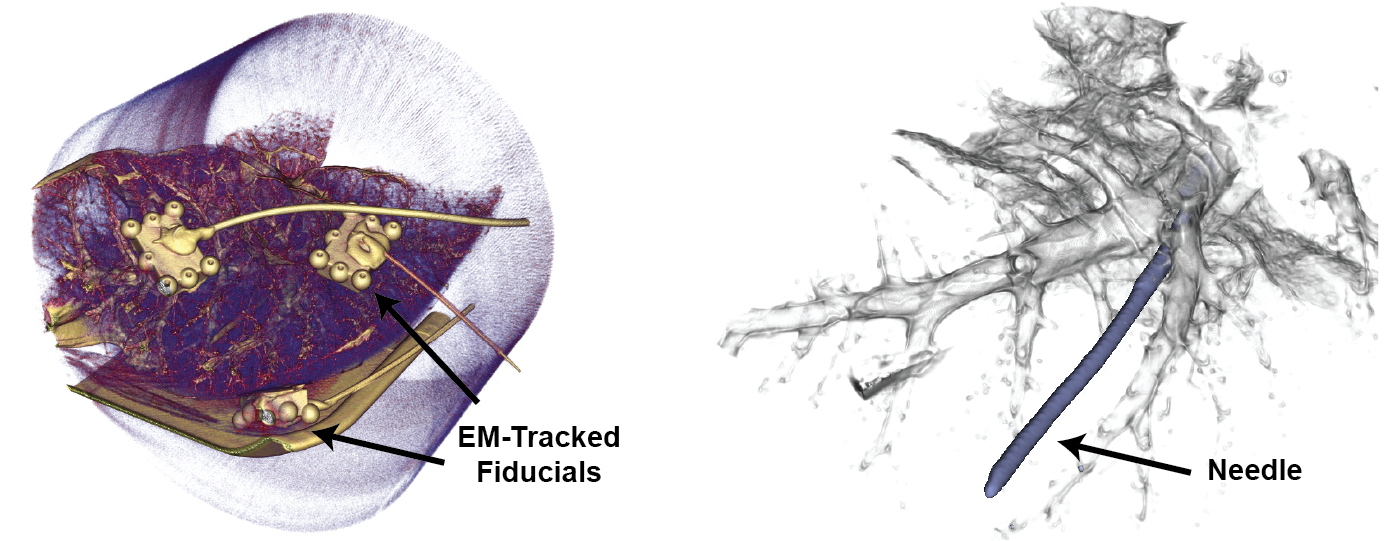}
    \caption{Rendered CT Scan Volume of the post-steered needle system using the learned estimation method. \textit{Left:} The fiducials are shown in the scan; each contains a 6DOF EM sensor and is glued to the surface of the lung, and is used for point-based registration of the CT frame to the EM tracker frame. \textit{Right:} The same thresholded scan showing the segmented needle deployed post-steer.
    }
    \vspace{-0.7cm}
    \label{fig:rendered_ct}
\end{figure*}

\textbf{Online Steering in \textit{Ex Vivo} Porcine Lung}
In addition to evaluating the estimator performance in gelatin and ovine brain, we performed a sequence of online steers in statically-inflated \emph{ex vivo} porcine lung (Animal Technologies, Inc.).
The \emph{ex vivo} lung was inflated and accessed using an 8.0 mm endotracheal tube (Smiths Medical ASD, Inc.). We placed custom 3D printed (Formlabs, Inc.) pre-calibrated fiducials on the lung surface with cyanoacrylate glue and used them for point-based registration of the EM tracker frame to the CT frame \cite{Fitzpatrick1998}. A preoperative CT scan was taken using a mobile ENT cone-beam CT scanner (xCAT Xoran Technologies).
We loaded the scan into 3D Slicer \cite{Slicer2014} and manually segmented the fiducial points (sphere centers) in the CT frame. We then manually thresholded the reconstructed CT data (\unit[0.4]{mm} isotropic voxel size) to yield a segmentation of the lung anatomy, see Fig.~\ref{fig:rendered_ct}.
We then registered the segmented anatomy in the CT frame to the EM tracker frame using the EM-tracked fiducials mounted on the lung.
We used this registered segmentation to inform the piercing sites and the intended target points.

Using a clinical bronchoscope, we navigated down to several airways in the left lower lobe and pierced through the airway wall in each trial using a piercing stylet---a \unit[0.9]{mm} OD superelastic Nitinol tube sharpened to a needle point. We inserted a nitinol tube, \unit[1.5]{mm} OD, \unit[1.3]{mm} ID, over the stylet to hold the opening in the airway wall. The piercing stylet was removed and exchanged for the steerable needle.  

\begin{figure*}[t]
    \centering
    \includegraphics[width=\textwidth]{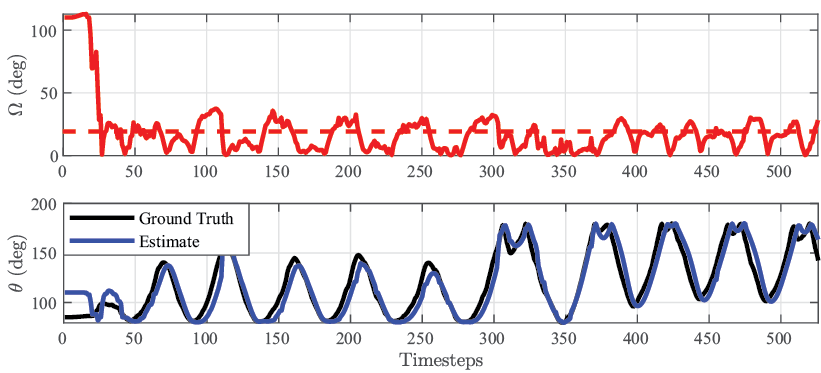}
    \vspace{-0.7cm}
    \caption{Time series of the estimate tracking the needle roll angle $\theta$ with angular error $\Omega$ from a steering trial in ex-vivo porcine lung using the learned estimator. The dashed line shows the mean angular error over the steer. 
    }
    \vspace{-0.5cm}
    \label{fig:lung_timeseries}
\end{figure*}

After loading the needle into the piercing site, we visualized the needle's trumpet-shaped reachable workspace with respect to the segmentation to identify feasible target points for each steer that were collision-free with respect to blood vessels and other surrounding airways. 
This target point, specified in the CT scanner RAS (right, anterior, superior) coordinates, was transformed into the EM tracker frame using the registration transform acquired from the fiducials.
We fed the target point to the sliding mode controller, and the needle was steered to the target with the estimated roll angle as input to the controller.
A total of 8 trials were executed, with collision-free steers in 2/8, 1 with each method. 
We determined the steer was collision-free by inspection of a CT scan of the post-steered needle, prior to retracting the needle back to its starting pose.
It is important to note that while we attempted to pick target points in the CT scan that were not visibly in-collision with other parts of the airway and large vessels, we do not consider obstacle avoidance in this work and the steers were performed without consideration for obstacles en-route to the target point. As such, we limit this evaluation to the collision-free trials.
In the collision-free trial of the learned estimator, the steer was accurate, achieving a targeting error of \unit[0.46]{mm} and \unit[19.0]{$^\circ$} average angular error of the estimate, as measured in the EM tracker frame, not accounting for registration error (see Fig.~\ref{fig:lung_timeseries}). 
Conversely, the EKF performed poorly, consistent with our prior experimental results in the other tissues, with a target error of \unit[17.6]{mm} and \unit[156.7]{$^\circ$} average angular error. 
A rendered CT scan of the post-steered needle is shown for the learned estimator method trial in Fig.~\ref{fig:rendered_ct}, right.
\vspace*{-0.2cm}

\section{Experimental Insights and Future Work}
\vspace*{-0.3cm}
In this work, we leveraged and validated the performance of a learned estimator for needle steering, paving the way for other learned methods in this application.

Our results show that a learning method can accurately estimate the roll angle of a long, flexible steerable needle in the presence of torsional compliance and unmodeled frictional effects.

Additionally, we show that a network trained on tissue simulant extrapolates to actual tissue in the form of \emph{ex vivo} ovine brain and porcine lung. It will be a subject of future research to investigate the application to other tissues such as liver and kidney, relevant tissues with applications for steerable needles.

In the \emph{ex vivo} lung experiments, the learned estimator performed well, though there were steers which collided with anatomy. This further elucidates the need for better registration methods, improved segmentation algorithms, and planned trajectories that consider these obstacles, instead of point targets. Future research will leverage our method when executing trajectories generated via motion planning in these environments.

Overall, we show that our learning-based method outperforms a model-based observer whose model does not capture the significant torsional dynamics. We demonstrate that neural network-based estimation can result in effective tracking of unsensed state variables, enabling accurate steering to point targets in gelatin, ovine brain, and porcine lung. 

\textbf{Acknowledgments}
This research was supported in part by the National Institutes of Health under award R01EB024864.

\vspace*{-0.3cm}

\bibliographystyle{IEEEtran}
\bibliography{literature}

\end{document}